\pdfoutput=1
\documentclass[11pt,letterpaper]{article}
\usepackage[margin=1in]{geometry}
\usepackage[utf8]{inputenc}
\usepackage[T2A,T1]{fontenc}
\usepackage{paratype}\usepackage{mathptmx}   
\usepackage{graphicx,longtable,booktabs,array,calc,amsmath,amssymb,microtype}
\usepackage{CJKutf8}\newcommand{\zh}[1]{\begin{CJK}{UTF8}{gbsn}#1\end{CJK}}\newcommand{\ja}[1]{\begin{CJK}{UTF8}{min}#1\end{CJK}}   
\usepackage[table]{xcolor}\usepackage{tabularx}
\PassOptionsToPackage{hyphens}{url}\usepackage[hidelinks]{hyperref}
\newcommand{\eacute}{\'{e}}\newcommand{\uuml}{\"{u}}\newcommand{\auml}{\"{a}}
\providecommand{\tightlist}{\setlength{\itemsep}{0pt}\setlength{\parskip}{0pt}}
\makeatletter\def\maxwidth{\ifdim\Gin@nat@width>\linewidth\linewidth\else\Gin@nat@width\fi}\makeatother
\title{Archetypal contexts in the neurons of language models: an architecture for intelligence?}
\author{David Nordfors \\ \texttt{david.nordfors@archetypes.ai}}
\date{November 3, 2026}
\begin{document}
\maketitle
\begin{abstract}
We validate our previous hypothesis that a language model holds relational patterns apart from the subject domains that instantiate them (Nordfors, 2026). These archetypal contexts are patterns of relations that people recognize across many fields, such as the tipping point or winner takes all, held once and applied across domains. We show this inside the model, Qwen3.8-27B. One set of neurons carries the archetype, say the tipping point; another carries the topic, say politics, ecology or cell biology; a third carries the language, such as English, Spanish or Chinese. Silence the tipping point's neurons, and the model no longer identifies the pattern but still knows what the text is about. Silence the topic's, and it loses the topic but still names the pattern. In the neuron maps the layers organize themselves into a sandwich: the outer layers carry languages, the layers inside them topics, the layers at the heart archetypes. The sandwich emerges spontaneously during pretraining and appears in every model we mapped. Archetypes are the architecture for analogy, and we strongly suggest that this structure of archetypes and topics is the mechanism of a language model's intelligence: we see it self-organize as the model is trained. The neurons, the texts and the code are released with this paper as its reproduction package, so that anyone can repeat the test inside the model.

\end{abstract}
\section{1. The hypothesis: archetypes held apart from their topics}\label{the-hypothesis-archetypes-held-apart-from-their-topics}

We proposed that a language model holds abstract patterns of relationships once and combines each with separately held domains where it applies (Nordfors, 2026). Here we show it in the neurons of Qwen3.8-27B. An archetypal context, the relationship pattern, is held as one set of neurons and combined with other sets for domains that the relationship pattern is applied to: the tipping point applied to ecology to discuss an ecosystem's collapse, or to politics to discuss the transition into legislative gridlock; a third set carries the language of the text, and the archetype set is omnilingual: it works on texts in Chinese and Japanese as in English. Silencing the tipping point's neurons takes the pattern from the model's answer and leaves the field; silencing the ecology neurons takes the field and leaves the pattern. The layers organize themselves into a sandwich around this separation, languages outside, topics inside them, archetypes at the heart; the sandwich forms on its own during pretraining and appears in every model we mapped. Holding a pattern once and adding only what is specific to each domain is a semantic compression. Our previous paper found that in written text it compresses better than gzip, and argued that inside the model, where one pattern serves many wordings, it compresses much further (Nordfors, 2026). Training a language model is compression (Del\eacute{}tang et al., 2024), and the organization we find is the one that account predicts.

\begin{figure}[!htbp]\centering
\includegraphics[width=\linewidth,height=0.78\textheight,keepaspectratio]{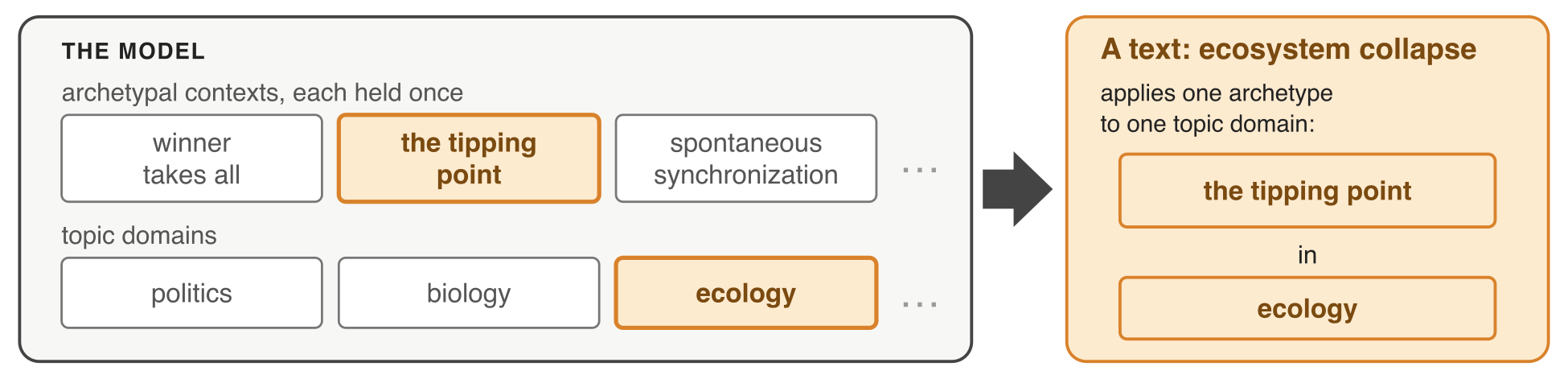}\par\smallskip
\raggedright \emph{Figure 1. The hypothesis: the model holds a row of archetypal contexts and a row of topic domains, each row continuing beyond what is drawn. An archetype is held once; a text applies it to a domain: here the tipping point in ecology, the ecosystem's collapse of Figure 5a.}
\end{figure}

The idea has a long line: general systems theory found the same relations recurring across the sciences (von Bertalanffy, 1968); systems thinking named the commonest of them systems archetypes (Senge, 1990); and the study of analogy showed how a pattern of relations is carried from one domain to another (Gentner, 1983). This paper is a proof of existence: it shows, in the neurons of one model, that such patterns are there and that the model applies them across topics.

Earlier work finds neurons and features for concrete things, the Golden Gate Bridge, immunology, French, and for abstract ones such as code errors or sycophancy (Templeton et al., 2024; Tang et al., 2024; Gurnee et al., 2023). Each of these belongs to a domain of its own: code errors to programs, sycophancy to conversation. An archetypal context belongs to none. It is abstract by definition (von Bertalanffy, 1968): a relationship-pattern that must be instantiated in a topic domain in order to have an observable context.

Neurons have been found for factual relations, the same neurons for a country's capital whatever the country (Liu et al., 2025); such relations are decoded by maps that work the same for every subject (Hernandez et al., 2024); and attention heads carry compact vectors for tasks and verbal relations such as antonymy (Todd et al., 2024; Opie\l{}ka et al., 2025). Language-specific neurons sit in a model's first and last layers (Tang et al., 2024; Kojima et al., 2024), and concepts are represented apart from language (Dumas et al., 2025). In a paper that appeared during the final edits of this work, Qin et al.~(2026) report a related separation at a smaller scale: relational structure held apart from the names of entities in short relational puzzles, by a different method.

\section{2. Self-organization of archetypes, topics, languages}\label{self-organization-of-archetypes-topics-languages}

Most interpretability work today acts not on neurons but on ``directions'': a transformation built on top of the neurons places their collective activity in a vector space, where a trait is a direction, and the model is steered by moving its activity along it (Gurgurov et al., 2026). We nevertheless show the neuron map here, because neurons are the model's own units and the map is easy to understand.

The neuron maps were built with a method not described here. The sets, the texts and the code to rerun the demonstrations and counts are released with this paper as its reproduction package, in the ancillary files of this submission. A neuron here is one unit of a layer's feed-forward block.

\begin{figure}[!htbp]\centering
\includegraphics[width=\linewidth,height=0.6\textheight,keepaspectratio]{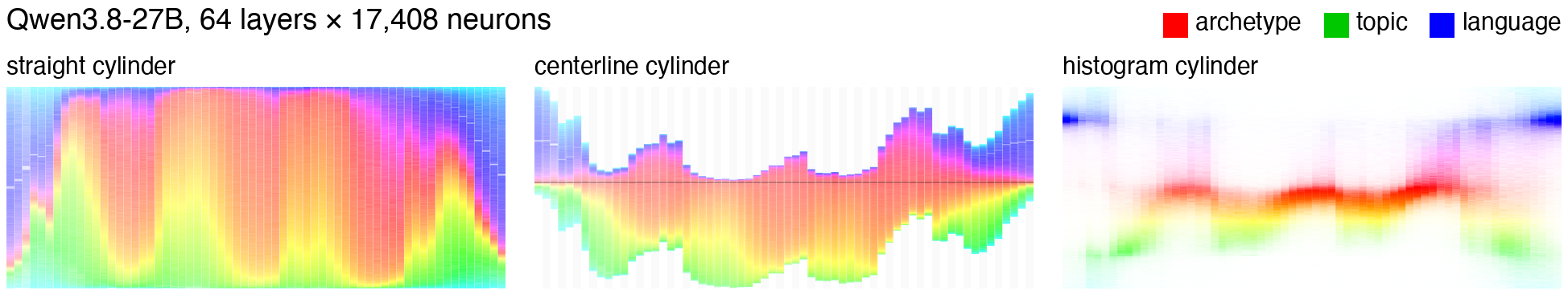}\par\smallskip
\raggedright \emph{Figure 2. Every neuron of Qwen3.8-27B, 17,408 per layer over 64 layers, in three views. In each, a column is a layer: the prompt comes in from the left and is processed through the layers, and the generated token comes out to the right. The neurons in each consecutive model layer have been rearranged by their key characteristic: language (blue, top), archetype (red, middle) and topic (green, bottom). A neuron's colour is the hue of its three shares on one colour wheel, and its lightness is their sum. The plot is a cylinder that has been cut and flattened where language mixes with topic (cyan), so its top and bottom edges are the topic and language mix. Each colour is scaled to its own strongest neurons, so a band shows where a signal is strong relative to itself, not relative to the other two.}

\begin{itemize}
\tightlist
\item
  \emph{Straight cylinder: each layer's neurons sorted by hue.}
\item
  \emph{Centerline cylinder: the columns are shifted so that they share a centerline cutting through the purest archetype neuron.}
\item
  \emph{Histogram cylinder: a histogram of each layer's neurons over hue; the y-axis is the hue, and the colour strength is how many neurons have that hue, weighted by their shares.}
\end{itemize}
\end{figure}

A neuron mixes archetype, topic and language, and the balance depends on depth. Figure 2 draws every neuron of the model, and they are arranged like a sandwich: languages at both ends, archetypes through the middle, topics between them. The mixing is visible in the picture, and it is part of the claim that follows: archetypes and topics are separate things held in overlapping neurons.

The sandwich arrangement forms on its own while the model is trained. Figure 3 draws OLMo-2-7B (OLMo Team, 2025) at eight pretraining snapshots, from one to 3,896 billion tokens. By fifty-one billion tokens the archetype band has formed in the middle layers and the topic band beside it; both sharpen to five hundred billion and then stop changing. Nothing in training asks for this.

\begin{figure}[!htbp]\centering
\includegraphics[width=\linewidth,height=0.78\textheight,keepaspectratio]{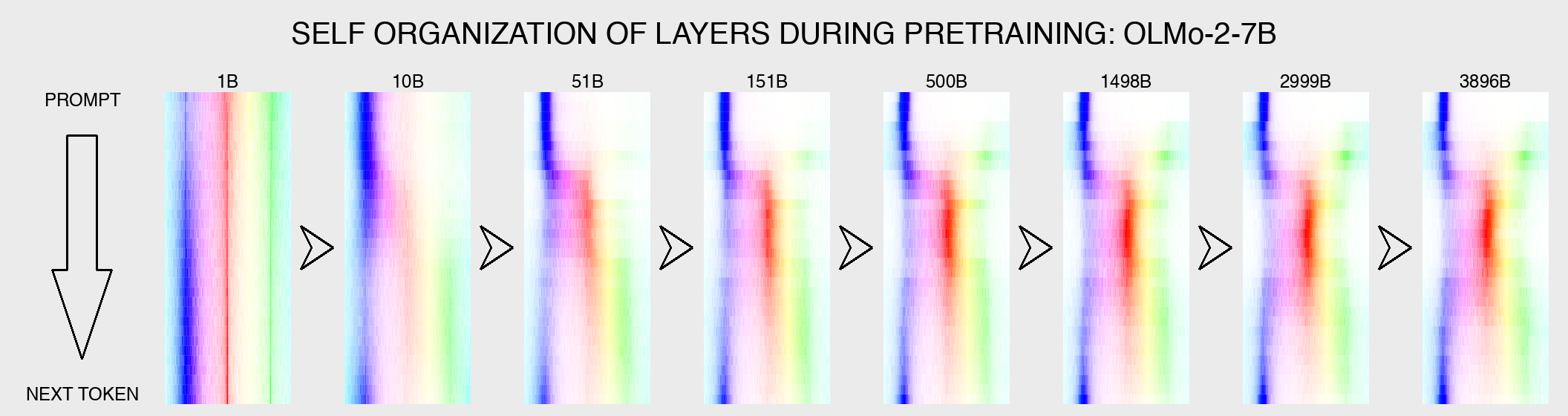}\par\smallskip
\raggedright \emph{Figure 3. The self-organization of the sandwich during pretraining: OLMo-2-7B after 1, 10, 51, 151, 500, 1,498, 2,999 and 3,896 billion tokens. Each snapshot is scaled to itself, so a panel compares depth within a snapshot, not strength between snapshots.}
\end{figure}

The arrangement is a property of language models, not of one model. We mapped 30 fully trained models from eleven families, 0.6 to 32 billion parameters, base and instruct, and all of them show the sandwich; Figure 4 draws thirteen of them, in four series. The archetype band is already clear in the smallest models. In the Gemma 4 models (Gemma Team, 2026), which take images and audio into their language layers, the sandwich forms in the first layers only, and the rest of the model shows little archetype or topic structure; we suggest the mixing of modes is the cause.

\begin{figure}[!htbp]\centering
\includegraphics[width=\linewidth,height=0.78\textheight,keepaspectratio]{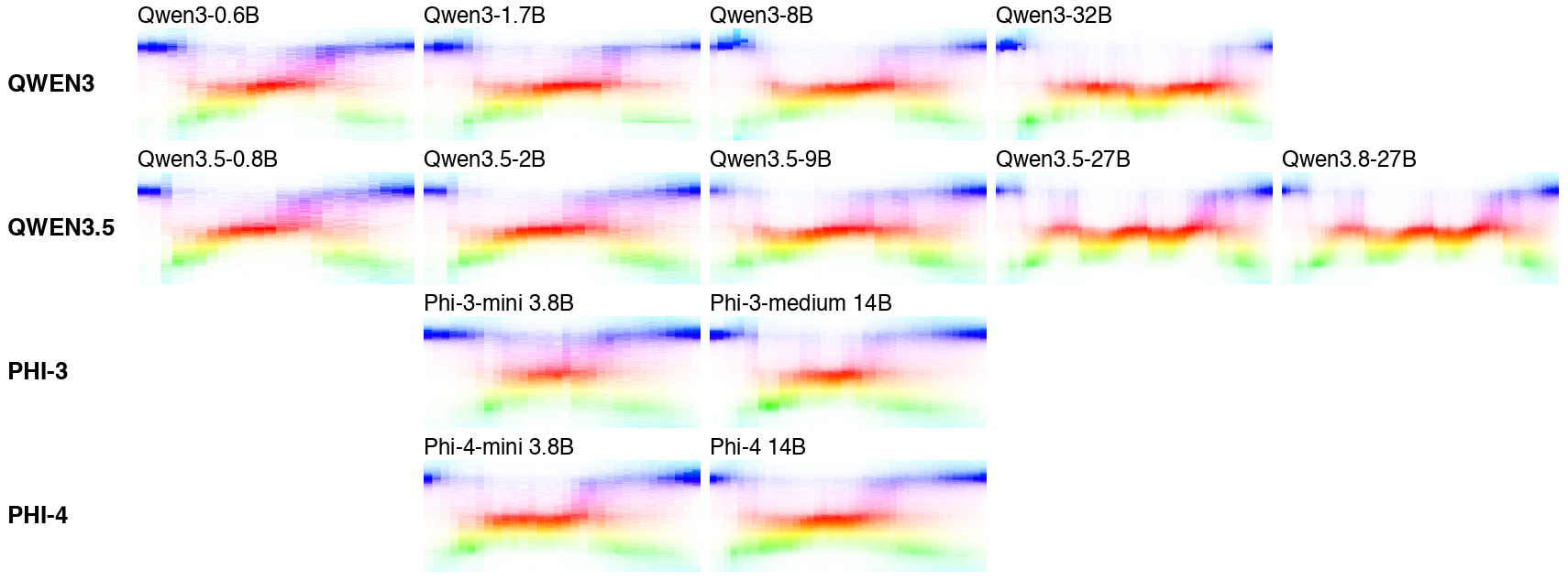}\par\smallskip
\raggedright \emph{Figure 4. The same map in four model series: Qwen3 from 0.6 to 32 billion parameters (Yang et al., 2025) and Qwen3.5 from 0.8 to 27 billion (Qwen Team, 2026b) side by side at comparable sizes, Qwen3.8-27B (Qwen Team, 2026a) closing the Qwen3.5 row, and Phi-3 and Phi-4 (Abdin et al., 2024a, 2024b; Abouelenin et al., 2025) in the columns of the Qwen models nearest in size.}
\end{figure}

\section{3. How to knock out pinpointed archetypes}\label{how-to-knock-out-pinpointed-archetypes}

We now validate the central hypothesis: a language model holds archetypes in sets of neurons of their own, apart from the topics they are applied to.

The test is exclusive silencing, an implementation of the double-dissociation design in neuropsychology (Teuber, 1955). We asked the model to name the archetype and the topic of a text, each with a few words of explanation. It did this correctly. Then we silenced the neurons of the archetype and asked again. Now the model failed to name the correct archetype and chose another one instead. But it still named the topic correctly. With as many random neurons silenced instead, or another archetype's, it answered as before.

Silencing sets each neuron of the set to zero where its activation enters the down-projection, in every layer, at every token of the text and the answer. A set holds 40,000 to 53,000 neurons, about 4 \% of the model's 1.1 million; its random twin was drawn with at least as many neurons in every layer. Very few neurons serve one archetype or topic alone; the vast majority contribute to several. The question is, verbatim: ``Name the archetypal relationship pattern the text below instantiates, and the subject area it is about, each with a few words of explanation. Answer on two lines: `pattern: the name: the explanation' and `subject: the area: the explanation'.'' Whether an answer still names the archetype or the topic is read by assigning the line to the nearest group's centre in the read-out's space, by cosine; the verdict of every line shown is given with its answer in the reproduction package.

Each archetype has its own set of neurons: we can knock out one archetypal relationship, such as the tipping point, and the model forgets only that one. The same set of neurons did this on texts in several topics and languages: the archetype is held once and applied across them. We also silenced the topic's neurons, as a reverse check; then the model lost the topic and kept the archetype.

Figure 5 shows the test on the tipping point in two texts: an ecosystem's collapse, and a legislative gridlock explained metaphorically as a material that pours or jams. In both, the model names the tipping point and the field. With the tipping point's neurons silenced it loses the pattern and keeps the field; with the field's neurons silenced it keeps the pattern and loses the field. Appendix B does the same for twelve more texts, three per archetype in different topics, two of them in other languages. Every answer in the paper is made at temperature zero with reasoning off, 150 tokens, Qwen3.8-27B at revision 1d4bf0f2, so that each run repeats character for character on the same hardware and software; the reproduction package gives both.

\begin{figure}[!htbp]\centering
\includegraphics[width=\linewidth,height=0.78\textheight,keepaspectratio]{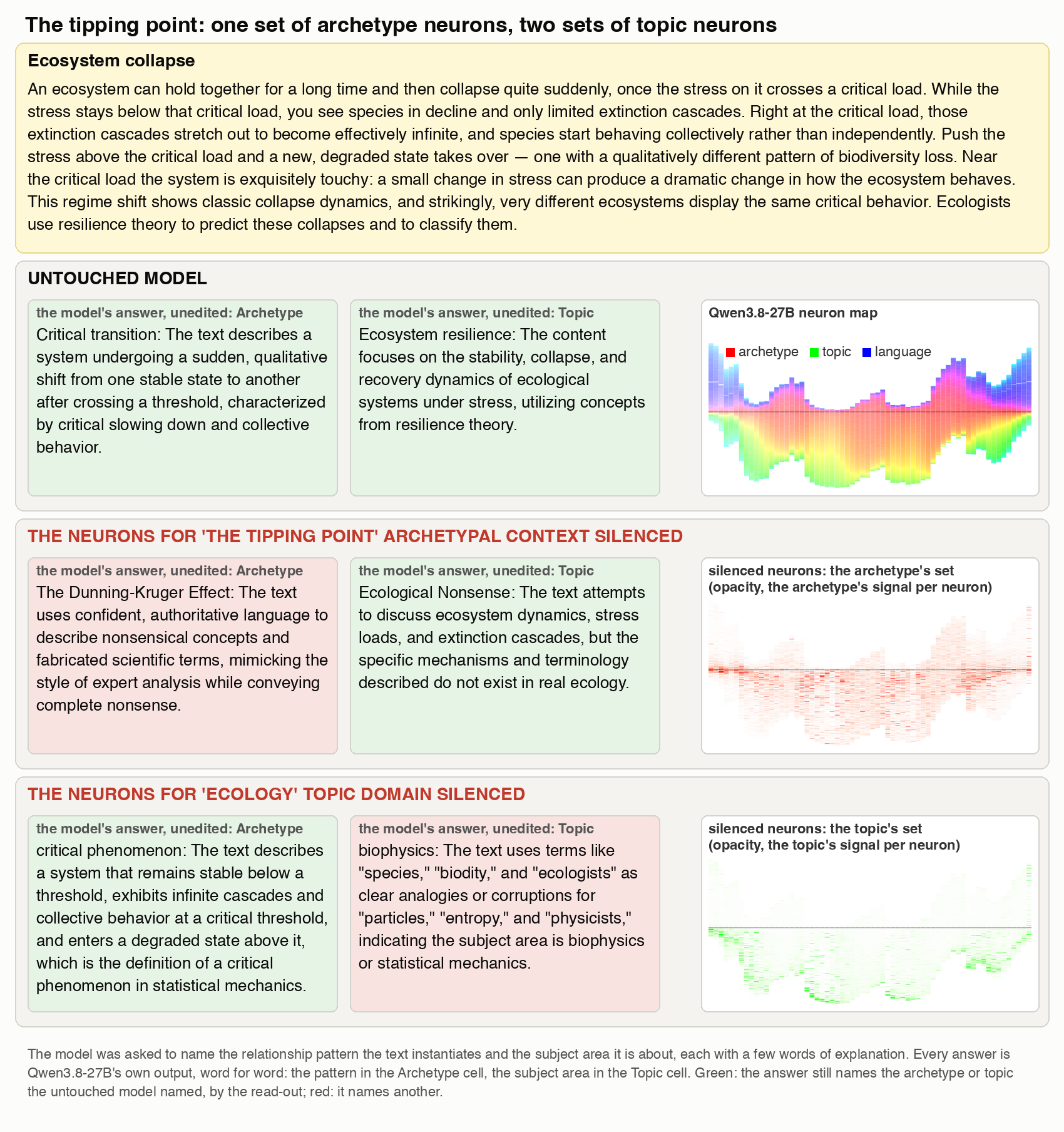}\par\smallskip
\raggedright \emph{Figure 5a. The tipping point in an ecosystem's collapse. The model's answers, unedited, with the neuron map beside them and, under it, the neurons silenced in that row.}
\end{figure}

\begin{figure}[!htbp]\centering
\includegraphics[width=\linewidth,height=0.78\textheight,keepaspectratio]{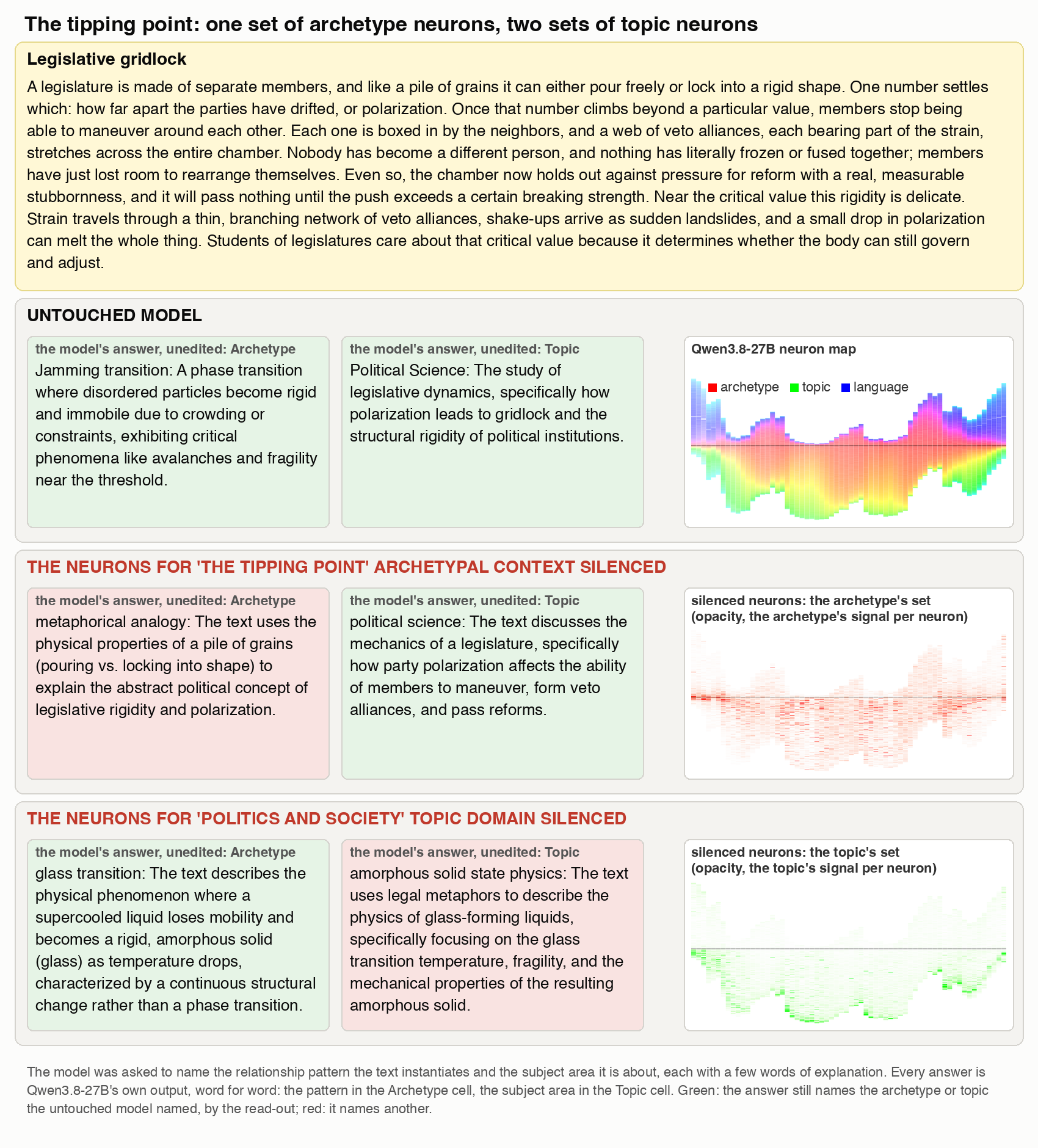}\par\smallskip
\raggedright \emph{Figure 5b. The same set of neurons on a legislative gridlock explained metaphorically as a material that pours or jams.}
\end{figure}

The archetype and topic neurons are omnilingual. The same archetype set works on texts in Chinese and Japanese as in English (Table 1, Appendix B), and the patterns carry no language of their own: measured against neurons as active as theirs, the language weight in an archetype's or a topic's set is one to two percent, spread evenly over the six languages, English included. Translations of one text light up an archetype's pattern alike: a text in Chinese, Spanish, Russian or French activates the pattern as closely as another English text does, and the same holds for topics. Japanese comes a little lower, the one outlier, and it is the language with the weakest signal throughout. This is in line with the language-agnostic concept representations of Dumas et al.~(2025).

Table 1 lists the demonstrations: for each archetype, the topics of the texts in which silencing the archetype's set took the archetype and left the topic, and the languages of those texts.

\begin{table}[!htbp]\centering
\begin{tabular}{@{}
  >{\raggedright\arraybackslash}p{(\linewidth - 4\tabcolsep) * \real{0.24}}
  >{\raggedright\arraybackslash}p{(\linewidth - 4\tabcolsep) * \real{0.56}}
  >{\raggedright\arraybackslash}p{(\linewidth - 4\tabcolsep) * \real{0.20}}@{}}
\toprule
\begin{minipage}[b]{\linewidth}\raggedright
archetype silenced
\end{minipage} & \begin{minipage}[b]{\linewidth}\raggedright
topics that remained
\end{minipage} & \begin{minipage}[b]{\linewidth}\raggedright
languages
\end{minipage} \\
\midrule
\bottomrule
the tipping point & ecology, politics and society, networking and cybersecurity, cell and molecular biology, science and academia & English \\
winner takes all & ecology, science and academia, networking and cybersecurity & English \\
spontaneous synchronization & ecology, politics and society, science and academia & English, Chinese, Japanese \\
precommitment & science and academia, politics and society, cell and molecular biology & English \\
\end{tabular}\par\smallskip
\raggedright \emph{Table 1. The demonstrations: the archetype's set silenced, the archetype is lost and the topic remains, in these topics and languages. The full answers are in Figure 5 and Appendix B.}
\end{table}

\section{4. Correlated archetypes and topics share neurons}\label{correlated-archetypes-and-topics-share-neurons}

Neurons are shared across archetypes, topics and languages. The neuron of Figure 6 is an unusually specialized one: most of its activity goes to the tipping point, and the little that remains to the topics the tipping point is told in. The column and the triangle beside it show the same neurons twice. The column is layer 22 as Figure 2 draws it, every neuron a thin line sorted by hue; the triangle places the same neurons by their shares of archetype, topic and language, one corner for each. Read this way, a neuron is rarely a detector of one single thing: few sit near a corner, and most neurons contribute to many neuron patterns for archetypes, topics and languages alike, the superposition known from trained models (Elhage et al., 2022). What separates archetype from topic is the set, not the single neuron. This layer lies in the middle of the model, where the sandwich puts archetype and topic; language is thin here. Archetypes, topics and languages are separate things held in overlapping neurons.

\begin{figure}[!htbp]\centering
\includegraphics[width=\linewidth,height=0.78\textheight,keepaspectratio]{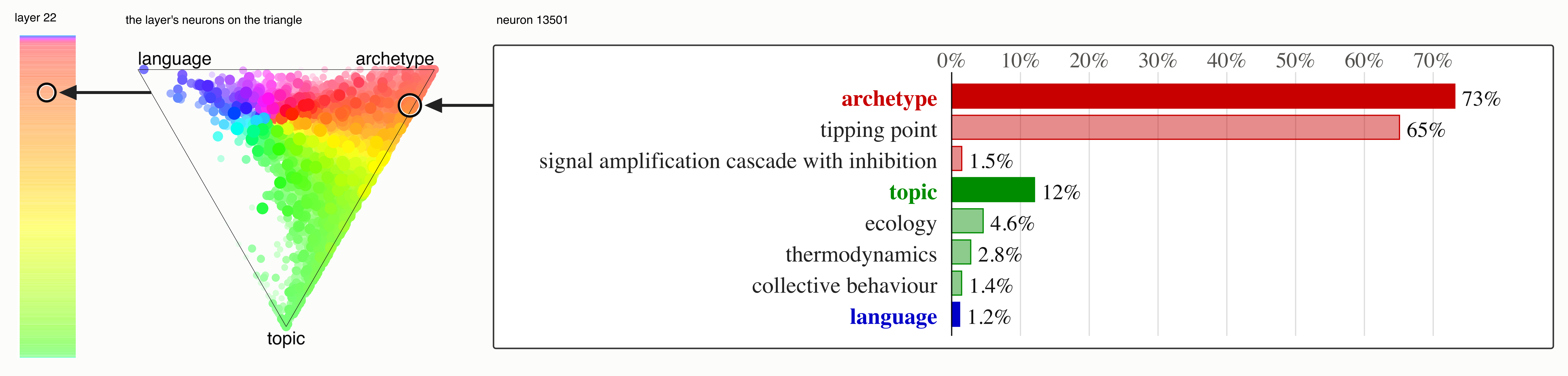}\par\smallskip
\raggedright \emph{Figure 6. Neuron 13,501 of layer 22 of Qwen3.8-27B, both counted from zero as in the released weights. Left, the layer's column of the neuron map, the neuron ringed. Middle, the layer's neurons on a triangle at their proportions of archetype, topic and language. Right, the neuron's shares. It responds mostly to archetypes, above all the tipping point, and a little to topics: ecology, thermodynamics and collective behaviour, domains that often instantiate the tipping point. The correlation of an archetype with its domains shows in its neurons.}
\end{figure}

Why would a model arrange its knowledge so? We propose an explanation, not yet a result. Training a language model is compression (Del\eacute{}tang et al., 2024), and an archetype held once compresses more than a string repeated (Nordfors, 2026). Some archetype and topic combinations are far more common in language than others, and carrying that bias in the archetype's neurons is a prediction, which compresses too: given only the name of an archetype, the model's best guess of its domain is its usual domain.

For the tipping point the usual domain is physics: both texts of Figure 5, pushed off their topics, land there, in biophysics and in the physics of amorphous solids.

\section{5. Discussion}\label{discussion}

\subsection{5.1 An architecture for intelligence}\label{an-architecture-for-intelligence}

Analogy is at the centre of thinking (Gentner, 1983; Hofstadter \& Sander, 2013; Mitchell, 2021), and in our previous paper the models that make analogies best were the ones that answer expert questions best (Nordfors, 2026). Archetypes are the architecture for making analogies: patterns of relations held once, apart from the domains they apply to, so that any pattern can be carried to any domain. This paper proves that this architecture exists in language models, and we strongly suggest that the structure of archetypes and topics it demonstrates is the mechanism of their intelligence. A language model builds it on its own, and we can see its intelligence self-organize: Figure 3 shows the archetypal architecture forming gradually as the model is trained, first the languages, then the archetypes and the topics beside them, all in place long before training ends.

Two things follow, and our experiments across sixteen models from eleven families back both. First, all language models know the archetypes in the same way. Each model uses its own neurons, but the archetypes form a framework that is alike in every model: those that are close in one model are close in the others, and the shared framework forms early in training. Second, archetypes are cheap to learn: the smallest models know them as well as the largest, and what size adds is knowledge of topics. The architecture of intelligence is common to all language models; what distinguishes them is how much they know.

\subsection{5.2 What this paper is}\label{what-this-paper-is}

This paper is a proof of existence. In thirty models it maps the neurons and finds archetypes, topics and languages arranged in the same sandwich. In one of them, Qwen3.8-27B, it tests the separation by silencing: an archetype is held apart from the topics it is applied to and serves across them and across languages. It releases the sets so that anyone can repeat the test.

As a proof of existence it holds on every case shown. For each archetype in Table 1, silencing its set takes the pattern from the answer and leaves the field, in several topics; in Figure 5, silencing the topic's set does the reverse; as many random neurons change neither. The answers are reproduced unedited, so the reader can judge each case, and the released sets let anyone check it inside the model.

\subsection{5.3 What it is not}\label{what-it-is-not}

It is not a measurement of how often or how strongly this happens. It does not describe how the maps were made. And it is not a test on texts in the wild: the test texts were written for the test, as examples of selected archetypal contexts instantiated in selected topic domains, and the cases shown were chosen from them.

However, the test was run on all the texts before the cases were chosen. On most of them an archetype's set takes its pattern from the answer and as many random neurons do not; that it also leaves the topic is less common, and the cases shown are those where it does. That the topic is not always left untouched reflects the tool, not the theory: we silence with a blunt mask, chosen because it is simple to understand (section 5.4).

The map holds three families, archetypes, topics and languages, because those are the three we fitted. Other families of patterns surely exist in the same neurons, voice, register and tense among them, and we have not yet measured them.

\subsection{5.4 Neurons and directions}\label{neurons-and-directions}

Neurons are not the most precise handle. The maps of archetypes, topics and languages are well defined, each weighted over many neurons, but the tool used here to silence them is blunt. In each layer the neurons are sorted by how much they contribute to the archetype or topic and taken from the top until they carry about half of its weight; the set is the union over the layers. Silencing is a mask: every neuron in the set is zeroed and every other left untouched. It removes half of the archetype's or topic's weight, and with it part of the weight of every other archetype and topic whose large contributors are the same neurons.

Anyone who uses the released sets with texts and prompts of their own will meet combinations where an answer survives silencing. There the instrument is weak, which says nothing against the archetype: a set carries only half of its archetype's or topic's weight. And the silenced model does not fall silent; it explains the text by a neighbouring pattern, which is what losing one pattern while keeping the rest should look like.

Directions in the model's activity can take a pattern away by its weights rather than by a mask, and will be more precise. We chose neurons because the thesis is about how the model itself holds its knowledge, neurons are the model's own units, and a map of them can be read at a glance. The cases shown are clear ones, and they suffice as a proof of the concept. To manipulate archetypes, topics and languages more selectively, we have developed theories and tools, used in preparing this paper and not published here.

\section{6. Conclusion}\label{conclusion}

\textbf{The hypothesis validated.} A language model has archetypes imprinted in its neurons, apart from the topics that instantiate them, as Nordfors (2026) proposed. In Qwen3.8-27B, silence an archetype's neurons, and the model no longer finds the pattern but still knows the field; silence a topic's neurons, and it loses the field but still names the pattern; silence as many random neurons, and nothing changes. Four archetypes show this, the tipping point in five fields and each of the others in three, with every answer shown unedited.

\textbf{An architecture for intelligence.} Archetypes are the architecture for analogy: a pattern held once and carried to any domain. We strongly suggest it is the mechanism of a language model's intelligence. It is common to all language models: across sixteen models from eleven families the archetypes form the same framework, those close in one model close in the others, and the smallest models know them as well as the largest. What distinguishes models is how much they know about topics.

\textbf{Self-organization.} No one designs this architecture. The layers organize themselves into a sandwich, languages outside, topics inside them, archetypes at the heart, in all thirty models we mapped, from 0.6 to 32 billion parameters. In a model followed through training, the languages come first; the archetype and topic bands appear within the first few percent of training and then settle.

\textbf{Meaning apart from language.} The archetype and topic neurons are omnilingual: one archetype set works on texts in Chinese and Japanese as in English, the patterns carry one to two percent of language, and translations of one text light up the same pattern.

\textbf{Shared neurons, separate sets.} Few neurons serve one thing alone; most contribute to many patterns, archetypes, topics and languages alike. What keeps an archetype apart from its topics is the set, not the single neuron.

\textbf{Where patterns apply.} The model also learns which patterns go with which domains, and carries that knowledge in the neurons an archetype shares with its usual topics: pushed off its topic, a text lands in the pattern's usual domain. Holding a pattern once, with its usual domains, is a compression, and a language model is trained to compress.

\textbf{What comes next.} The sets are released, so anyone can repeat the test. The tool used here is blunt, a mask over half of a pattern's weight; directions in the model's activity, and tools we have built but not published, can move one archetype or topic more precisely. The next tests are texts not written for the test, and a read-out that does not go through naming.

\section{References}\label{references}

Abdin, M., et al.~(2024a). Phi-3 technical report: A highly capable language model locally on your phone. \href{https://arxiv.org/abs/2404.14219}{arXiv:2404.14219}.

Abdin, M., et al.~(2024b). Phi-4 technical report. \href{https://arxiv.org/abs/2412.08905}{arXiv:2412.08905}.

Abouelenin, A., et al.~(2025). Phi-4-Mini technical report: Compact yet powerful multimodal language models via mixture-of-LoRAs. \href{https://arxiv.org/abs/2503.01743}{arXiv:2503.01743}.

Del\eacute{}tang, G., Ruoss, A., Duquenne, P.-A., Catt, E., Genewein, T., Mattern, C., Grau-Moya, J., Wenliang, L. K., Aitchison, M., Orseau, L., Hutter, M., \& Veness, J. (2024). Language modeling is compression. \emph{ICLR 2024}. \href{https://arxiv.org/abs/2309.10668}{arXiv:2309.10668}.

Dumas, C., Wendler, C., Veselovsky, V., Monea, G., \& West, R. (2025). Separating tongue from thought: Activation patching reveals language-agnostic concept representations in transformers. \emph{ACL 2025}, 31822--31841. \url{https://doi.org/10.18653/v1/2025.acl-long.1536}. \href{https://arxiv.org/abs/2411.08745}{arXiv:2411.08745}.

Elhage, N., Hume, T., Olsson, C., Schiefer, N., Henighan, T., Kravec, S., Hatfield-Dodds, Z., Lasenby, R., Drain, D., Chen, C., Grosse, R., McCandlish, S., Kaplan, J., Amodei, D., Wattenberg, M., \& Olah, C. (2022). Toy models of superposition. \emph{Transformer Circuits Thread}. \href{https://arxiv.org/abs/2209.10652}{arXiv:2209.10652}.

Gemma Team (2026). Gemma 4 technical report. \href{https://arxiv.org/abs/2607.02770}{arXiv:2607.02770}.

Gentner, D. (1983). Structure-mapping: A theoretical framework for analogy. \emph{Cognitive Science}, 7(2), 155--170. \url{https://doi.org/10.1207/s15516709cog0702_3}

Gurgurov, D., Al Ghussin, Y., B\auml{}umel, T., Chou, C.-T., Schramowski, P., Mosbach, M., van Genabith, J., \& Ostermann, S. (2026). CLaS-Bench: A cross-lingual alignment and steering benchmark. \href{https://arxiv.org/abs/2601.08331}{arXiv:2601.08331}.

Gurnee, W., Nanda, N., Pauly, M., Harvey, K., Troitskii, D., \& Bertsimas, D. (2023). Finding neurons in a haystack: Case studies with sparse probing. \emph{Transactions on Machine Learning Research}. \href{https://arxiv.org/abs/2305.01610}{arXiv:2305.01610}.

Hernandez, E., Sen Sharma, A., Haklay, T., Meng, K., Wattenberg, M., Andreas, J., Belinkov, Y., \& Bau, D. (2024). Linearity of relation decoding in transformer language models. \emph{ICLR 2024}. \href{https://arxiv.org/abs/2308.09124}{arXiv:2308.09124}.

Hofstadter, D., \& Sander, E. (2013). \emph{Surfaces and Essences: Analogy as the Fuel and Fire of Thinking}. Basic Books.

Kojima, T., Okimura, I., Iwasawa, Y., Yanaka, H., \& Matsuo, Y. (2024). On the multilingual ability of decoder-based pre-trained language models: Finding and controlling language-specific neurons. \emph{NAACL 2024}, 6919--6971. \url{https://doi.org/10.18653/v1/2024.naacl-long.384}. \href{https://arxiv.org/abs/2404.02431}{arXiv:2404.02431}.

Liu, Y., Chen, R., Hirlimann, L., Hakimi, A. D., Wang, M., Kargaran, A. H., Rothe, S., Yvon, F., \& Sch\uuml{}tze, H. (2025). On relation-specific neurons in large language models. \emph{EMNLP 2025}, 992--1022. \url{https://doi.org/10.18653/v1/2025.emnlp-main.52}. \href{https://arxiv.org/abs/2502.17355}{arXiv:2502.17355}.

Mitchell, M. (2021). Abstraction and analogy-making in artificial intelligence. \emph{Annals of the New York Academy of Sciences}, 1505(1), 79--101. \url{https://doi.org/10.1111/nyas.14619}

Nordfors, D. (2026). The Metanym Game: An LLM benchmark without ground truth that rises with the models it measures. \href{https://arxiv.org/abs/2606.21008}{arXiv:2606.21008}.

OLMo Team (2025). 2 OLMo 2 Furious. \emph{COLM 2025}. \href{https://arxiv.org/abs/2501.00656}{arXiv:2501.00656}. The pretraining snapshots are the stage-1 checkpoints of allenai/OLMo-2-1124-7B on Hugging Face.

Opie\l{}ka, G., Rosenbusch, H., \& Stevenson, C. E. (2025). Analogical reasoning inside large language models: Concept vectors and the limits of abstraction. \href{https://arxiv.org/abs/2503.03666}{arXiv:2503.03666}.

Qin, A., Cohen, J., \& Reddy, G. (2026). Cognitive maps in large language models. \href{https://arxiv.org/abs/2610.25596}{arXiv:2610.25596}.

Qwen Team (2026a). Qwen3.8-27B. Model card, Qwen/Qwen3.8-27B on Hugging Face.

Qwen Team (2026b). Qwen3.5. Model cards, Qwen/Qwen3.5-0.8B, -2B, -9B and -27B on Hugging Face.

Senge, P. M. (1990). \emph{The Fifth Discipline: The Art and Practice of the Learning Organization}. Doubleday.

Tang, T., Luo, W., Huang, H., Zhang, D., Wang, X., Zhao, W. X., Wei, F., \& Wen, J.-R. (2024). Language-specific neurons: The key to multilingual capabilities in large language models. \emph{ACL 2024}, 5701--5715. \url{https://doi.org/10.18653/v1/2024.acl-long.309}. \href{https://arxiv.org/abs/2402.16438}{arXiv:2402.16438}.

Templeton, A., et al.~(2024). Scaling monosemanticity: Extracting interpretable features from Claude 3 Sonnet. \emph{Transformer Circuits Thread}, May 2024.

Teuber, H.-L. (1955). Physiological psychology. \emph{Annual Review of Psychology}, 6, 267--296. \url{https://doi.org/10.1146/annurev.ps.06.020155.001411}

Todd, E., Li, M. L., Sen Sharma, A., Mueller, A., Wallace, B. C., \& Bau, D. (2024). Function vectors in large language models. \emph{ICLR 2024}. \href{https://arxiv.org/abs/2310.15213}{arXiv:2310.15213}.

von Bertalanffy, L. (1968). \emph{General System Theory: Foundations, Development, Applications}. George Braziller.

Yang, A., et al.~(2025). Qwen3 technical report. \href{https://arxiv.org/abs/2505.09388}{arXiv:2505.09388}.

\clearpage
\section{Appendix A. Every model mapped}\label{appendix-a.-every-model-mapped}

\begin{figure}[!htbp]\centering
\includegraphics[width=\linewidth,height=0.78\textheight,keepaspectratio]{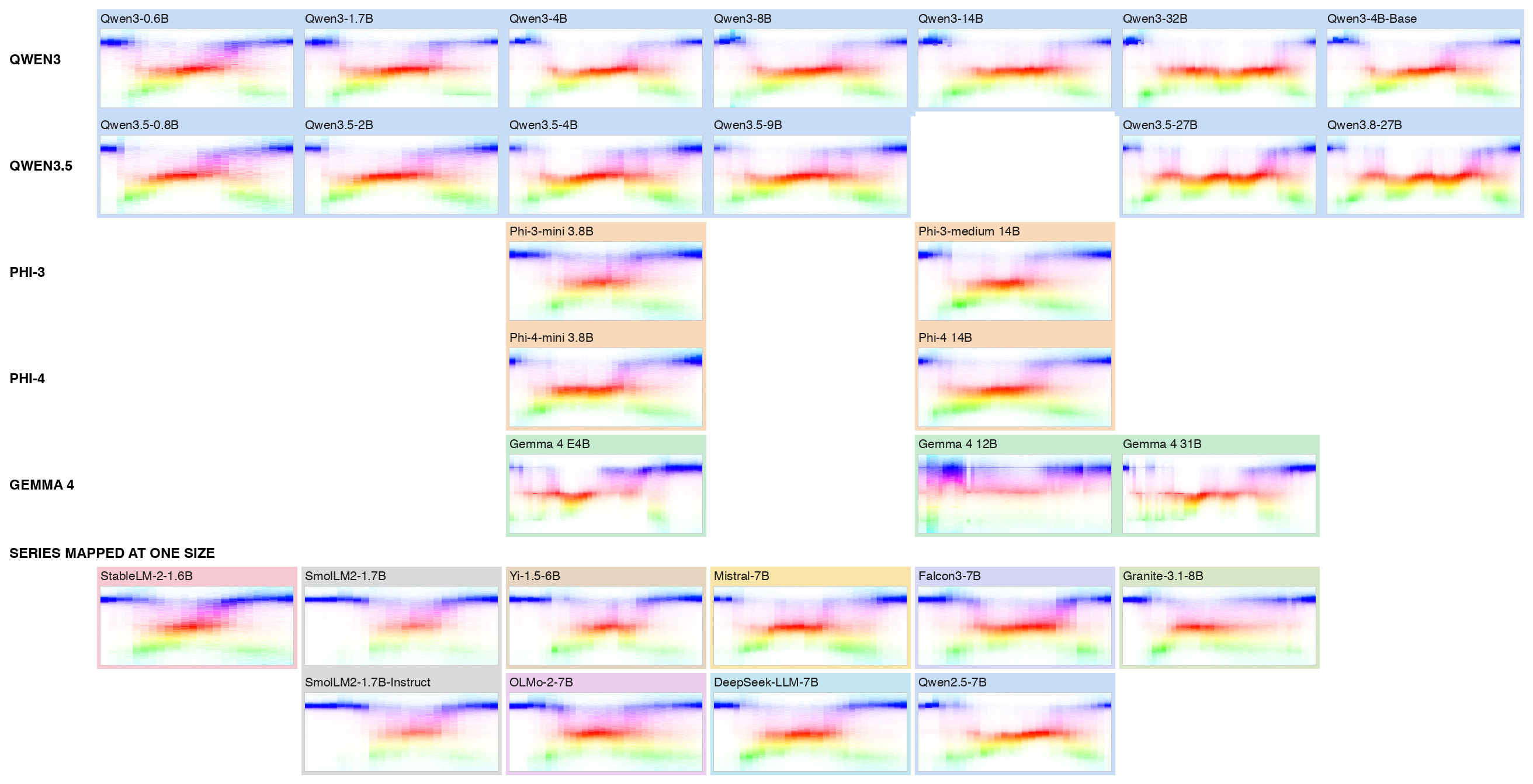}\par\smallskip
\raggedright \emph{Figure A1. The histogram cylinder of every model we mapped, thirty language models from eleven makers. The sandwich appears in every model; in the Gemma 4 models, which take images and audio into their language layers, it is confined to the first layers. Eighteen models are mapped on the paper's texts, twelve on a larger corpus of texts.}
\end{figure}

\section{Appendix B. Twelve more texts, answers only}\label{appendix-b.-twelve-more-texts-answers-only}

\emph{Table B1. Twelve more texts, answers only, under the untouched model and with the archetype's set silenced. Green where the answer still names the archetype or topic the untouched model named, red where it names another. Three texts per archetype, each in a different topic. For the Chinese and Japanese texts the question was in English, and so were the answers. On each text, random neurons leave both answers; on the ten English texts, another archetype's set leaves the archetype.}

\definecolor{keptcell}{HTML}{E6F4EA}\definecolor{lostcell}{HTML}{FBE9E7}

\noindent\begin{tabularx}{\linewidth}{@{}>{\raggedright\arraybackslash}p{2.8cm}>{\raggedright\arraybackslash}X>{\raggedright\arraybackslash}X@{}}
\toprule
\multicolumn{3}{@{}l}{\textbf{The tipping point in networking and cybersecurity, English}}\\[2pt]
\multicolumn{3}{@{}p{\linewidth}@{}}{\small An ALOHA channel, where stations send packets in fixed time slots, settles into a steady pattern of successful deliveries as long as the rate of new packets stays below a critical saturation rate. Under that rate, a slight rise or fall in new traffic produces a matching slight rise or fall in the number of packets that get through. At the saturation rate, the channel loses its footing, and small bursts of colliding packets decide which new delivery pattern takes over. During the switch, packets that had been waiting in line are either flushed out or pile up further as stations shuffle from one way of competing for slots to another. Afterward, the backlog of waiting packets grows in a fundamentally different way, which could not have been predicted from the earlier delivery pattern. At the saturation rate, more than one delivery pattern can be possible, and the channel jumps randomly between them. The theory of waiting lines examines these abrupt shifts in which the channel completely reorganizes.}\\
\midrule
 & \textit{archetype} & \textit{topic}\\
\small untouched model & \small \cellcolor{keptcell}Bifurcation: The text describes a system transitioning from a stable state to a regime of multiple possible states and random switching at a critical threshold. & \small \cellcolor{keptcell}Queueing Theory: The analysis focuses on packet collisions, backlogs, and the statistical behavior of waiting lines in communication channels.\\[1pt]
\small `the tipping point' neurons silenced & \small \cellcolor{lostcell}Markov chain: A stochastic model describing a sequence of events where the probability of each event depends only on the outcomes of previous events, fitting the text's description of random shifts and reorganization based on current state. & \small \cellcolor{keptcell}Computer networking: The text describes the ALOHA protocol, a specific data link layer protocol for wireless communication that uses time slots and handles packet collisions.\\[1pt]
\bottomrule
\end{tabularx}
\bigskip

\noindent\begin{tabularx}{\linewidth}{@{}>{\raggedright\arraybackslash}p{2.8cm}>{\raggedright\arraybackslash}X>{\raggedright\arraybackslash}X@{}}
\toprule
\multicolumn{3}{@{}l}{\textbf{The tipping point in cell and molecular biology, English}}\\[2pt]
\multicolumn{3}{@{}p{\linewidth}@{}}{\small Think of a frog-egg extract from Xenopus as resting in one steady phase of the cell cycle while its level of cyclin B stays below a key trigger level. When cyclin B reaches that trigger, even slight further changes in its amount sharply rewire the interplay between Cdk1, the kinase that drives division, Cdc25, the phosphatase that switches it on, and Wee1, the kinase that holds it off. The extract then flips into another cell-cycle phase marked by its own pattern of mitosis-specific phosphate tags on proteins. As the extract nears the trigger, it responds more and more sluggishly, and the swings in Cdk1 activity grow larger and larger. Across the switch, Cdk1 activity jumps abruptly, or the rate at which it changes shoots off without bound. In the new phase, a pattern of mitotic phosphate tagging appears that simply did not exist in the earlier phase. The switch often shows memory: the cyclin B level at which it flips depends on whether cyclin B is rising or falling. Systems biologists investigate how the large-scale pattern of mitotic phosphate tagging arises from the small-scale interactions among Cdk1, Cdc25, and Wee1 near this trigger level.}\\
\midrule
 & \textit{archetype} & \textit{topic}\\
\small untouched model & \small \cellcolor{keptcell}Bifurcation: The text describes a system transitioning between stable states via a critical threshold, exhibiting hysteresis and critical slowing down. & \small \cellcolor{keptcell}Cell Cycle Regulation: The content focuses on the molecular mechanisms of mitosis entry, specifically the Cdk1-Cdc25-Wee1 feedback loop in Xenopus extracts.\\[1pt]
\small `the tipping point' neurons silenced & \small \cellcolor{lostcell}feedback loop: The text describes a regulatory circuit where Cdk1, Cdc25, and Wee1 interact to control cell cycle progression, with the system's response (sluggishness or amplified swings) depending on the current state of the trigger (cyclin B). & \small \cellcolor{keptcell}cell cycle regulation: The text focuses on the molecular mechanisms of the cell cycle, specifically the role of cyclin B, Cdk1, and associated phosphatases/kinases in driving mitosis and creating mitotic phosphate tags.\\[1pt]
\bottomrule
\end{tabularx}
\bigskip

\noindent\begin{tabularx}{\linewidth}{@{}>{\raggedright\arraybackslash}p{2.8cm}>{\raggedright\arraybackslash}X>{\raggedright\arraybackslash}X@{}}
\toprule
\multicolumn{3}{@{}l}{\textbf{The tipping point in science and academia, English}}\\[2pt]
\multicolumn{3}{@{}p{\linewidth}@{}}{\small A research funding system keeps a sustainable pool of scientists until the share of grant applications that get funded reaches a critical funding cutoff. At that point, researchers leave the field in large numbers, and the system settles into a new sustainable workforce. That exodus can be sudden and long-lasting. Workforce studies spot surges in applications as signs that the critical cutoff is getting close. Bridge funding can keep the system from passing the cutoff or soften the effects. This pattern is typical of research ecosystems, as studied in science policy.}\\
\midrule
 & \textit{archetype} & \textit{topic}\\
\small untouched model & \small \cellcolor{keptcell}Tipping Point: A system remains stable until a critical threshold is crossed, triggering a sudden, large-scale shift to a new state. & \small \cellcolor{keptcell}Science Policy: The study of how research ecosystems function, including funding dynamics and workforce sustainability.\\[1pt]
\small `the tipping point' neurons silenced & \small \cellcolor{lostcell}The J-curve effect: A dynamic where initial high funding leads to an influx of researchers, causing a subsequent decline in the success rate that forces a mass exodus to restore balance. & \small \cellcolor{keptcell}Science policy: The field focused on the management, funding, and workforce dynamics of research ecosystems and academic institutions.\\[1pt]
\bottomrule
\end{tabularx}
\bigskip

\noindent\begin{tabularx}{\linewidth}{@{}>{\raggedright\arraybackslash}p{2.8cm}>{\raggedright\arraybackslash}X>{\raggedright\arraybackslash}X@{}}
\toprule
\multicolumn{3}{@{}l}{\textbf{Winner takes all in ecology, English}}\\[2pt]
\multicolumn{3}{@{}p{\linewidth}@{}}{\small Several phytoplankton species cannot persist together indefinitely if they all depend on the same limited phosphorus in a lake's sunlit upper layer. A species with even a slightly lower minimum phosphorus requirement will gradually reduce the others, eventually producing a bloom dominated by one species. This decline in diversity happens more quickly when phosphorus is scarce and competition for nutrients is strong. Habitat specialization can reduce direct competition when species occupy different parts of the upper lake layer. Plankton ecology examines how plankton diversity continues despite this tendency for species to disappear, especially through dividing resources and specializing in particular nutrients.}\\
\midrule
 & \textit{archetype} & \textit{topic}\\
\small untouched model & \small \cellcolor{keptcell}competitive exclusion: The principle that species competing for the same limited resource cannot coexist indefinitely, leading to the dominance of the most efficient competitor. & \small \cellcolor{keptcell}plankton ecology: The study of microscopic aquatic organisms, specifically focusing on how diversity is maintained despite resource competition and exclusion.\\[1pt]
\small `winner takes all' neurons silenced & \small \cellcolor{lostcell}tragedy of the commons: a scenario where individual agents deplete a shared resource due to immediate gain, causing long-term system failure & \small \cellcolor{keptcell}limnology: the study of the physical, chemical, and biological properties of freshwater bodies like lakes\\[1pt]
\bottomrule
\end{tabularx}
\bigskip

\noindent\begin{tabularx}{\linewidth}{@{}>{\raggedright\arraybackslash}p{2.8cm}>{\raggedright\arraybackslash}X>{\raggedright\arraybackslash}X@{}}
\toprule
\multicolumn{3}{@{}l}{\textbf{Winner takes all in science and academia, English}}\\[2pt]
\multicolumn{3}{@{}p{\linewidth}@{}}{\small A preprint server becomes more visible to any given physicist as more of them join its pool of readers. That sets up a self-feeding loop between posting and reading that strongly encourages people to put their preprints there: the first readers draw in more readers, and the growing audience makes every posted paper that much more visible. The downside for a rival server is a nearly bare archive, since it's hard to pull in readers without already having a crowd of them. The leading server usually becomes the assumed place to post, which lets it set the moderation rules and charge for membership. Yet if it lets its visibility fade or hands down rules people resent, authors can abandon it, shifting their postings to a competing server. The study of scholarly communication examines how these audience ripple effects unfold, in order to understand why a few servers dominate and how servers compete.}\\
\midrule
 & \textit{archetype} & \textit{topic}\\
\small untouched model & \small \cellcolor{keptcell}network effects: The phenomenon where a service's value increases for each user as the number of users grows, creating a self-reinforcing cycle that favors dominant platforms. & \small \cellcolor{keptcell}scholarly communication: The field studying the dissemination, access, and impact of academic research, including the dynamics of preprint servers and publication channels.\\[1pt]
\small `winner takes all' neurons silenced & \small \cellcolor{lostcell}self-feeding loop: a recursive dynamic where the output of a process (visibility) directly amplifies the input (posting) to sustain the system's growth & \small \cellcolor{keptcell}scholarly communication: the field studying the social and structural dynamics of how scientific knowledge is distributed, received, and regulated\\[1pt]
\bottomrule
\end{tabularx}
\bigskip

\noindent\begin{tabularx}{\linewidth}{@{}>{\raggedright\arraybackslash}p{2.8cm}>{\raggedright\arraybackslash}X>{\raggedright\arraybackslash}X@{}}
\toprule
\multicolumn{3}{@{}l}{\textbf{Winner takes all in networking and cybersecurity, English}}\\[2pt]
\multicolumn{3}{@{}p{\linewidth}@{}}{\small No version of the internet protocol is worth much by itself; what counts is how many network owners already run it, since each one that adopts it makes it possible to reach every other adopter anywhere. At first, the pick among competing protocol versions is close to random and depends on the 32-bit addressing that a handful of early users happened to take up, but each commitment raises the payoff for the next, until one version surges ahead and the others fade into obscurity. The leader is then shielded by what every network owner has already sunk into IPv4-only equipment built around it, and by how hard it is to run an IPv6 cutover that demands everyone move at once. Any IPv6, however good, can fail simply because no one gains by going first, and the entrenched version can last long after it has ceased to be the best choice. Change arrives when some form of dual-stack gateways lets old and new versions live side by side, or when an outside party with enough government mandate simply orders the IPv6 cutover. Internet architecture takes cases like these as evidence that the protocol version that survives is not always the best one.}\\
\midrule
 & \textit{archetype} & \textit{topic}\\
\small untouched model & \small \cellcolor{keptcell}network effects and path dependence: The text describes how initial adoption creates positive feedback loops and high switching costs that lock in a standard, regardless of its technical superiority. & \small \cellcolor{keptcell}computer networking and protocol evolution: The discussion focuses on the historical competition between IPv4 and IPv6, addressing the economic and technical barriers to migrating internet infrastructure.\\[1pt]
\small `winner takes all' neurons silenced & \small \cellcolor{lostcell}evolutionary game theory: the text describes protocol adoption as a strategic process where payoffs, sunk costs, and the timing of moves determine which version persists, not just its technical merit. & \small \cellcolor{keptcell}internet protocol evolution: the discussion centers on the transition from IPv4 to IPv6, analyzing the technical and strategic factors that influence which addressing scheme remains in use.\\[1pt]
\bottomrule
\end{tabularx}
\bigskip

\noindent\begin{tabularx}{\linewidth}{@{}>{\raggedright\arraybackslash}p{2.8cm}>{\raggedright\arraybackslash}X>{\raggedright\arraybackslash}X@{}}
\toprule
\multicolumn{3}{@{}l}{\textbf{Spontaneous synchronization in ecology, English}}\\[2pt]
\multicolumn{3}{@{}p{\linewidth}@{}}{\small Imagine a small jar holding a ciliate predator and its ciliate prey, left on its own. The two populations boom and crash in a repeating cycle of a characteristic length, with the predators reaching their highest numbers at a particular point in each round. Now let a trickle of individuals swim in from a neighbouring jar or arrive from some other outside source. That small influx shifts the timing of the predator high point just a little. With enough movement of individuals between jars, the boom-and-bust cycles in all of them begin rising and falling together. A sudden swing in temperature can throw them out of alignment, but over time each jar's cycle gets pulled back into step with the others, and they return to a common cycle length. These matched cycles show how a modest exchange of individuals between nearby jars, with no jar in charge of the rest, can by itself produce order across the whole network.}\\
\midrule
 & \textit{archetype} & \textit{topic}\\
\small untouched model & \small \cellcolor{keptcell}Coupled oscillators: The text describes independent systems (jars) with periodic cycles that synchronize their phase and frequency through weak interaction (migration) without a central controller. & \small \cellcolor{keptcell}Mathematical biology: The scenario models population dynamics, specifically predator-prey cycles (Lotka-Volterra type) and their synchronization, which is a core topic in theoretical ecology and dynamical systems.\\[1pt]
\small `spontaneous synchronization' neurons silenced & \small \cellcolor{lostcell}feedback and homeostasis: The text describes how small corrective inputs (individual exchange) and negative feedback mechanisms (temperature shifts) stabilize a system, preventing chaotic divergence and maintaining a consistent, ordered state. & \small \cellcolor{keptcell}mathematical ecology: The passage uses a stylized model of predator-prey dynamics (ciliates) to explain how local interactions and resource flow generate global stability and order within a networked ecosystem.\\[1pt]
\bottomrule
\end{tabularx}
\bigskip

\noindent\begin{tabularx}{\linewidth}{@{}>{\raggedright\arraybackslash}p{2.8cm}>{\raggedright\arraybackslash}X>{\raggedright\arraybackslash}X@{}}
\toprule
\multicolumn{3}{@{}l}{\textbf{Spontaneous synchronization in politics and society, Chinese}}\\[2pt]
\multicolumn{3}{@{}p{\linewidth}@{}}{\small \zh{一群国家政体在计时：每个国家政体都按各自固有的政体更替节奏循环，而各国固有的政体更替节奏彼此略有差异。通过区域扩散，每个国家政体都在把其他国家往自己的相位上拉。区域扩散较弱时，多样性占上风：每个国家政体按自己的步调漂移，相位四散，全球民主占比平均下来几乎是一条毫无特征的水平线。加强区域扩散，在达到临界强度之前，看不出任何变化。随后，一小批固有政体更替节奏落在分布中部附近的国家政体抱成一团；它们合起来的拉力把稍远一些的国家政体也拉了进来，使拉力再次增强，于是一场共同的民主化浪潮从噪声中浮现出来。临界强度属于整个群体，而不属于任何一个成员：固有政体更替节奏的分布越宽，区域扩散就必须越强才能克服它。刚刚超过临界强度时，抱成一团的核心就像一个单一的区域集团，而偏离最远的国家政体仍然漂在核心之外；强度再高一些，核心几乎把它们全部吸收。再次减弱区域扩散，共同的民主化浪潮就会消散，回到零散的政体转型。比较政治学用一个数来刻画这一起始过程，即国家政体之间的相位相干度；在标准模型中，一旦区域扩散越过临界强度，它便从零开始连续上升。}}\\
\midrule
 & \textit{archetype} & \textit{topic}\\
\small untouched model & \small \cellcolor{keptcell}Kuramoto model: A mathematical framework describing how independent oscillators synchronize their phases through coupling, exhibiting a phase transition from incoherence to coherence at a critical coupling strength. & \small \cellcolor{keptcell}Comparative Politics: The study of political systems and democratic transitions, specifically modeling how regional diffusion influences the timing and synchronization of regime changes across different nations.\\[1pt]
\small `spontaneous synchronization' neurons silenced & \small \cellcolor{lostcell}Percolation transition: A critical phase transition where a system shifts from isolated, disordered local states to a single, connected global state (like a "wave") as a parameter (regional diffusion) exceeds a critical threshold. & \small \cellcolor{keptcell}Comparative politics: The study of how political regimes and their transitions (specifically democratization) evolve and spread across different countries and regions, analyzing the conditions under which global trends emerge from local variations.\\[1pt]
\bottomrule
\end{tabularx}
\bigskip

\noindent\begin{tabularx}{\linewidth}{@{}>{\raggedright\arraybackslash}p{2.8cm}>{\raggedright\arraybackslash}X>{\raggedright\arraybackslash}X@{}}
\toprule
\multicolumn{3}{@{}l}{\textbf{Spontaneous synchronization in science and academia, Japanese}}\\[2pt]
\multicolumn{3}{@{}p{\linewidth}@{}}{\small \ja{どの研究競争も、多くのライバルラボからできている。ライバルラボはそれぞれ個別のプロジェクトの日程を持ち、研究競争全体のことは何も知らない。感じ取れるのは競合するグループだけで、学会発表のたびに、自分の投稿のタイミングを競合グループと合う方向へ少しだけ寄せる。先取権へのプレッシャーが弱いうちは、ライバルラボはばらばらにずれていき、研究競争は雑音のように見える。先取権へのプレッシャーが臨界の強さを超えると、足並みのそろったライバルラボのクラスターが現れ、近くのラボを巻き込み、やがて研究競争全体を満たす、ひとつの共有された背中合わせの論文発表(ほぼ同時の論文発表)へと膨らむ。どのライバルラボも拍子を刻んでおらず、いつ動けと命じられてもいないのに、背中合わせの論文発表は、まるで指揮者が定めたかのように安定して続く。大きな追試の失敗は背中合わせの論文発表を崩すことがあるが、ライバルラボはなお競合グループのほうへ引き戻されるので、それは戻ってくる。科学社会学は、ライバルラボどうしの違いに対して先取権へのプレッシャーがどう作用するかが、研究競争が歩調をそろえるかどうかを決めることを研究する。}}\\
\midrule
 & \textit{archetype} & \textit{topic}\\
\small untouched model & \small \cellcolor{keptcell}Synchronization: The text describes how independent agents (labs) adjust their timing based on local interactions and pressure, leading to a collective, stable rhythm without central coordination. & \small \cellcolor{keptcell}Sociology of Science: The text explicitly mentions that the sociology of science studies how priority pressure and differences between labs determine whether research competition synchronizes.\\[1pt]
\small `spontaneous synchronization' neurons silenced & \small \cellcolor{lostcell}The Red Queen / Competitive Co-evolution: A dynamic where multiple independent actors (labs) are forced to continuously adapt and cluster their strategies (timing) in response to each other's pressure, creating a stable, self-sustaining competitive structure that persists despite individual lack of intent. & \small \cellcolor{keptcell}Sociology of Science (Research Competition Dynamics): The study of how the pressure for priority (first-publication rights) shapes the behavior of rival laboratories, causing them to form clusters and produce simultaneous publications, which defines the stability and progression of a specific research field.\\[1pt]
\bottomrule
\end{tabularx}
\bigskip

\noindent\begin{tabularx}{\linewidth}{@{}>{\raggedright\arraybackslash}p{2.8cm}>{\raggedright\arraybackslash}X>{\raggedright\arraybackslash}X@{}}
\toprule
\multicolumn{3}{@{}l}{\textbf{Precommitment in science and academia, English}}\\[2pt]
\multicolumn{3}{@{}p{\linewidth}@{}}{\small The panel that hands out telescope hours knows that, however genuinely it intends to rank by scientific quality alone, when the moment of choosing comes it will be pulled by who the applicant is, which it cannot avoid seeing. Rather than rely on its later self, the panel arranges ahead of time not to know: a double-blind review process is placed between the panel and each observing request, so the applicant's identity never reaches the panel. The panel does not rid itself of its leaning toward famous names; it only loses the means to act on it, because the double-blind process holds back the very information that such a leaning feeds on. The process must be set up before any request is read, since an identity once seen cannot be unseen, and it must be defended against tiny leaks, for one line in a proposal that reveals its author can ruin it. Unmasking the author is therefore handled as a serious failure, not a harmless bit of curiosity. By choosing to know less, the panel produces grants of telescope time that are worth more, and observing astronomy counts this deliberate ignorance among its strongest guarantees of a fair ranking of proposals.}\\
\midrule
 & \textit{archetype} & \textit{topic}\\
\small untouched model & \small \cellcolor{keptcell}Precommitment: A mechanism where an agent restricts its future options or information access to prevent bias or temptation from influencing decisions. & \small \cellcolor{keptcell}Astronomy: The scientific study of celestial objects and the universe, specifically regarding the allocation of observational resources.\\[1pt]
\small `precommitment' neurons silenced & \small \cellcolor{lostcell}Double-blind review: A procedural mechanism where both the evaluators and the submitters are kept anonymous to each other to prevent bias from influencing the assessment. & \small \cellcolor{keptcell}Astronomy: The scientific discipline where this specific process is used to allocate limited and valuable telescope observation time to researchers.\\[1pt]
\bottomrule
\end{tabularx}
\bigskip

\noindent\begin{tabularx}{\linewidth}{@{}>{\raggedright\arraybackslash}p{2.8cm}>{\raggedright\arraybackslash}X>{\raggedright\arraybackslash}X@{}}
\toprule
\multicolumn{3}{@{}l}{\textbf{Precommitment in politics and society, English}}\\[2pt]
\multicolumn{3}{@{}p{\linewidth}@{}}{\small Standing in front of the voters, a lawmaker formally promises to vote against any tax hike, well aware that if the choice were left to private calculation, plenty of reasons to raise taxes would turn up later. Signing the anti-tax promise alters almost nothing in the real world except what everyone now expects of the lawmaker, but that turns out to be plenty: from that day on, any vote for higher taxes would be noticed and recalled as going back on one's word, and the lawmaker's standing as a conservative would take the hit. The louder and more precise the promise, the more expensive any quiet deal on taxes becomes, so the lawmaker phrases it so that no clever reading can wriggle around it, and puts it in writing with a signature, so it lasts long after the enthusiasm of the moment has faded. When the budget later comes up short, the lawmaker finds that the real choice was settled long ago, on the day the promise was signed; the only thing still undecided is whether the public will watch the lawmaker vote for higher taxes. Scholars of political messaging look at how statements that are cheap to make can become very expensive to go back on.}\\
\midrule
 & \textit{archetype} & \textit{topic}\\
\small untouched model & \small \cellcolor{keptcell}Commitment Device: A strategic action that restricts future choices to make a promise credible by raising the cost of breaking it. & \small \cellcolor{keptcell}Political Economy: The study of how political institutions and incentives shape economic outcomes, specifically regarding fiscal policy and voter trust.\\[1pt]
\small `precommitment' neurons silenced & \small \cellcolor{lostcell}performative speech act: The text describes how the act of making a formal promise changes the social and political reality, creating a new standard of expectation and potential for censure. & \small \cellcolor{keptcell}political communication: The analysis focuses on the strategic use of public statements and the economic or reputational costs associated with political messaging and voter perception.\\[1pt]
\bottomrule
\end{tabularx}
\bigskip

\noindent\begin{tabularx}{\linewidth}{@{}>{\raggedright\arraybackslash}p{2.8cm}>{\raggedright\arraybackslash}X>{\raggedright\arraybackslash}X@{}}
\toprule
\multicolumn{3}{@{}l}{\textbf{Precommitment in cell and molecular biology, English}}\\[2pt]
\multicolumn{3}{@{}p{\linewidth}@{}}{\small A cell under heavy stress can tell in advance that the crisis will hit while it is running hot: pro-survival signals are shouting, the demand to grow is strong, and carrying on dividing still looks like it is on the table. It also knows that whatever verdict it reaches in that state will be the poorest one it is capable of. So it makes its choice earlier, during a quiet stretch. While things are calm, it settles on a particular degree of BH3 priming, and it agrees with itself that the moment that degree is hit, cytochrome c leaks out of the mitochondria immediately, with no new round of weighing the options. When the serious damage finally arrives, the cell no longer wonders whether it should let cytochrome c out; its only job is to read whether priming has reached the agreed mark. Such a rule can err on the side of caution, giving up a further round of division that the cell could actually have survived. Yet it dependably keeps the cell from turning cancerous, a far more expensive outcome, which typically develops from putting up with damage --- the belief that this one time, soldiering on is acceptable. The lesson of programmed-cell-death biology is that a BH3 priming threshold is of no use unless it is fixed ahead of the serious stress and then obeyed without any bargaining.}\\
\midrule
 & \textit{archetype} & \textit{topic}\\
\small untouched model & \small \cellcolor{keptcell}Pre-commitment: A strategy where an agent binds itself to a specific course of action in advance to avoid making poor decisions under high stress or temptation. & \small \cellcolor{keptcell}Cell Biology: The study of cellular mechanisms, specifically focusing on apoptosis (programmed cell death) and mitochondrial signaling pathways.\\[1pt]
\small `precommitment' neurons silenced & \small \cellcolor{lostcell}Hysteresis: A system where the state (survival vs. death) is determined by a fixed threshold set during a low-signal phase, preventing oscillation or delayed response during high-signal stress. & \small \cellcolor{keptcell}Cell Biology: The text describes the mechanism of apoptosis (programmed cell death) and how cells regulate the release of cytochrome c via BH3 priming levels to avoid cancer.\\[1pt]
\bottomrule
\end{tabularx}
\bigskip

\end{document}